\documentclass[letterpaper, 10 pt, conference]{ieeeconf}  

\IEEEoverridecommandlockouts                              

\usepackage{graphicx} 
\usepackage{subfigure}	
\usepackage{diagbox}
\usepackage{multirow}
\usepackage{amsmath} 
\usepackage{amsfonts,amssymb}
\usepackage{color}
\usepackage{cite}
\usepackage{aecompl}

\title{\LARGE \bf
HiTPR: Hierarchical Transformer for Place Recognition in Point Cloud
}

\author{Zhixing Hou$^{1}$, Yan Yan$^{1}$, Chengzhong Xu$^{2}$ and Hui Kong$^{\dag3}$
\thanks{$^{\dag}$ Corresponding author.}
\thanks{$^{1}$ Zhixing Hou and Yan Yan are with the School of Computer Science and Engineering, Nanjing University of Science and Technology, Nanjing, China. {\tt\small E-mail: \{hzx, yyan\}@njust.edu.cn}}
\thanks{$^{2}$ Chengzhong Xu is with the Department of Computer Science, University of Macau, Macau, China. {\tt\small E-mail: czxu@um.edu.mo}}
\thanks{$^{3}$ Hui Kong is with the Faculty of Science and Technology, University of Macau, Macau, China. {\tt\small E-mail: huikong@um.edu.mo}}
}

\begin{document}

\maketitle

	
	\begin{abstract}
		Place recognition or loop closure detection is one of the core components in a full SLAM system.	In this paper, aiming at strengthening the relevancy of local neighboring points and the contextual dependency among global points simultaneously, we investigate the exploitation of transformer-based network for feature extraction, and propose a \textbf{Hi}erarchical \textbf{T}ransformer for \textbf{P}lace \textbf{R}ecognition (HiTPR). The HiTPR consists of four major parts: point cell generation, short-range transformer (SRT), long-range transformer (LRT) and global descriptor aggregation. Specifically, the point cloud is initially divided into a sequence of small cells by down-sampling and nearest neighbors searching. In the SRT, we extract the local feature for each point cell. While in the LRT, we build the global dependency  among all of the point cells in the whole point cloud. Experiments on several standard benchmarks demonstrate the superiority of the HiTPR in terms of average recall rate, achieving 93.71\% at top 1\% and 86.63\% at top 1 on the Oxford RobotCar dataset for example.
    \end{abstract}
	

	\section{Introduction}
	
	In robotics or autonomous driving domain, Simultaneous Localization and Mapping (SLAM) system plays key roles on high-precision navigation. Generally, a SLAM system usually suffers from drift pose errors over time due to intrinsic sensor noise. These drift errors could be corrected by incorporating sensing information taken from places that have been visited before, a.k.a., loop closure detection, which requires algorithms that are able to recognize revisited places.
	
	\begin{figure*}[t]
		\centering
		\includegraphics[width=0.7\textwidth]{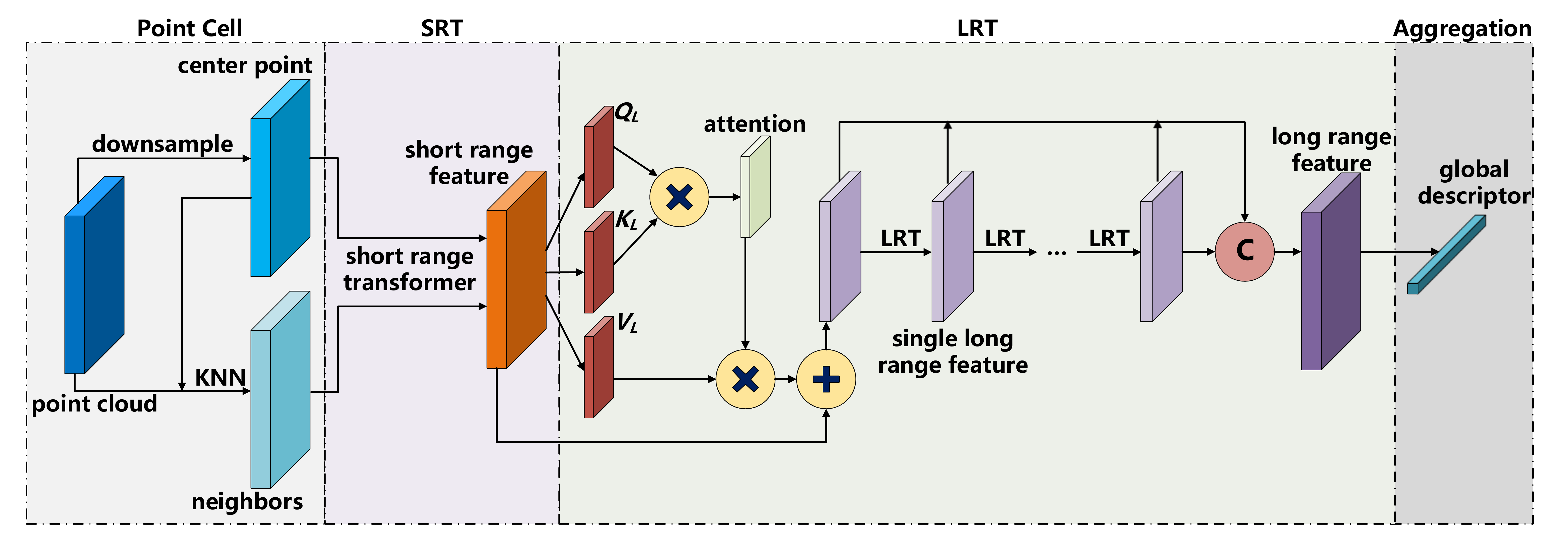}
		\caption{Our network architecture consists of point cell generation, short range transformer, long range transformer and aggregation modules.}
		\label{fig:long-range-transformer}
	\end{figure*}
	
	The revisited place can trigger the back-end optimization to reduce the accumulated drift error over time once 
	it is recognized. Conventionally, 
	place recognition or loop closure detection problem has been investigated by exploiting visual information from cameras, and has achieved significant advances 
	in the past twenty years \cite{VPR-Bench}. However, performance of visual methods is significantly affected by illumination variation, camera's field of view and viewing orientation.
	
	Thanks to the rapid development of the LiDAR technology, place recognition or loop closure detection in point cloud has drawn more and more attention recently \cite{Wang2020LiDARIF, segmatch2017, Uy2018PointNetVLADDP, Liu-LPD-Net, du2020dh3d, vid2021locus, Xia2020SOENetAS, Zhou2021NDTTransformerL3,lcdnet, Komorowski_2021_WACV,chen2020overlapnet,he2016m2dp,kim2018scan}. In contrast to cameras, LiDAR has a 360-degree field of view, and LiDAR rays are not easily affected by lighting changes. A point cloud obtained by modern LiDAR sensors can contain more than hundreds of thousands of 3D points of the world, which can capture quite decent details of surrounding geometric structure of places. Therefore, it make senses to exploit LiDAR sensor for 
	place recognition irrespective of the time of day and even weather conditions as well.


	Unfortunately, existing solutions for 
	place recognition with 3D LiDAR data are neither robust nor fast enough to meet the demand of real-world SLAM applications. For example, PointNetVLAD \cite{Uy2018PointNetVLADDP} and PCAN \cite{Zhang2019PCAN3A} treat point cloud as a sequence of un-ordered independent points, both of which ignore the relevancy of local structural points in the point cloud and get a relatively low accuracy. SOE-Net \cite{Xia2020SOENetAS} integrates local information for a point by adopting three different directional convolutions to the cube where the point lies. The convolution kernels are fixed once trained, even though the contents of the point clouds are various. To capture local information, some works focus on adding additional handcrafted features to the input of the network. LPD-Net \cite{Liu-LPD-Net} extracts 10 types of artificially designed features offline and feeds them into the network in parallel with the original points. NDT-Transformer \cite{Zhou2021NDTTransformerL3} has to discretize the raw point cloud into regular grid map and represents the grid cells with Normal Distribution Transform in advance. The pre-processing procedure in the two methods are inapplicable for real-time loop closure detection in SLAM. Apart from these, most of the methods utilizing NetVLAD \cite{NetVlad} as the aggregation module to gather the global information are memory-intensive and inefficient.
	
	To address the issues mentioned above, we propose a novel \textsl{transformer}  architecture named \textbf{HiTPR} (\textbf{Hi}erarchical \textbf{T}ransformer for \textbf{P}lace \textbf{R}ecognition)
	for large scale 
	place recognition problem. The transformer's self-attention mechanism has already made significant achievement in NLP (natural language processing) \cite{tranformer, devlin-etal-2019-bert, brown2020language} and CV (computer vision) \cite{chen2020generative,detr,dosovitskiy2020image,ipt} domains. 
	Different from very recent transformer models for 
	place recognition problem, e.g., the SOE-Net \cite{Xia2020SOENetAS} and NDT-Transformer \cite{Zhou2021NDTTransformerL3}, where the transformer mechanism is explored only by building long range context dependencies after point feature extraction, we fully investigate the transformer's capabilities on both extracting local features and gathering global contextual relationships among all points in the point cloud as well.
	
	Compared to the regular convolution operator with fixed kernels, the self-attention in transformer is adaptive to the content of the input point cloud, thus resulting in more robust and generative features. Besides, because the shape of point cloud has the characteristic of invariance to permutation of points and can be considered as a collection of unordered discrete points, it is innately suitable to be processed with transformer.
	In transformer \cite{tranformer}, the input data is transformed to three distinct embedding vectors habitually called $Query$, $Key$ and $Value$ in NLP, respectively. We follow the naming convention in this paper. The output of the transformer is the weighted sum of the $Value$ and the weights which are derived from the relation function between $Query$ and $Key$.
	
	In our work, the point cloud is firstly processed online, split into a sequence of small cells by down-sampling and nearest neighbors searching. Each cell is a small set of the points containing a sample center point and its neighbors that may exhibit varying shape. 
	Then, in the short range transformer (SRT), 
	we extract the local features of the sampled center points within the cells where the sampled points are located. Sampling and short range transformer are beneficial to model the relations of the neighboring points in Euclidean space and save the computational resources. However, the sampling and short range transformer operations break the inherent structural information in the point cloud. Therefore, we also introduce a long range transformer (LRT),
	where the global relation among the point cells in the whole point cloud is encoded by stacking the transformer blocks with the residual connection. Short range transformer and long range transformer are cascaded in our network. 

	\begin{figure*}[t]
		\centering
		\includegraphics[width=0.64\textwidth]{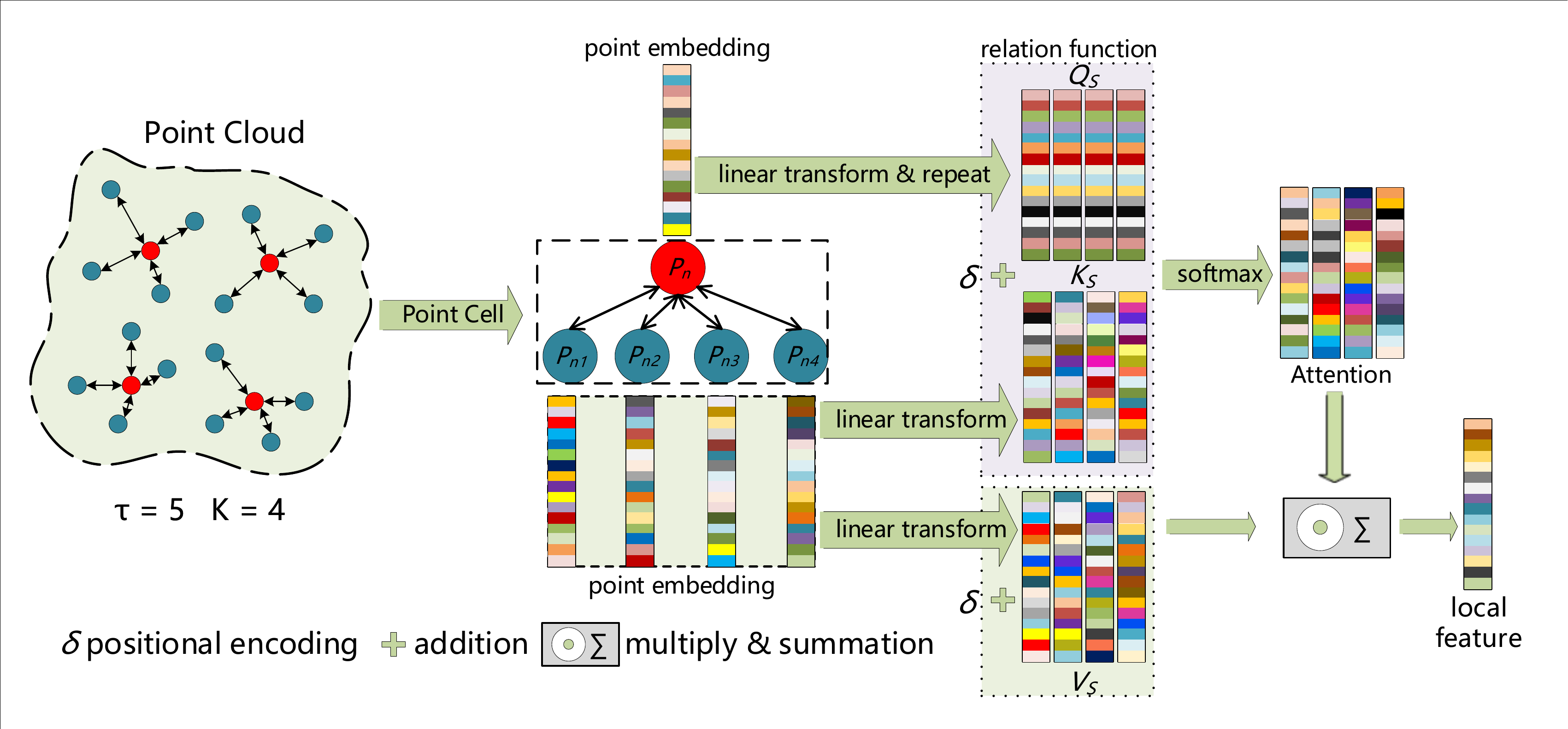}
		\caption{The short range transformer architecture for each point cell. Red points denote the sampled center points to be transformed to $Q_S$ vectors while aqua green points denote the neighboring points to be transformed to $K_S$ and $V_S$ vectors. The sampling rate $\tau$ and the number $K$ of the neighbors can be adjusted in practice.}
		\label{fig:short-range-transformer}
	\end{figure*}

	\section{Related works}
	\label{sec:Related_work}
	
	\subsection{Handcrafted Descriptors for PR and LCD}
	Handcrafted descriptors are applied to loop closure detection at the earliest. ESF (Ensemble of Shape Functions) \cite{ESF} introduces a global shape descriptor to describe distance, angle and area distributions on the surface of a point cloud. PFH (Persistent Feature Histogram) \cite{point-hist} and FPFH (Fast Point Feature Histogram) \cite{fpfh} extract a local geometry feature around each point and treat the feature histogram as the descriptor to align two point clouds with overlap views. SegMatch \cite{segmatch2017} recognizes  places by matching the corresponding 3D segments but doesn't rely on the perfect segmentation. M2dp \cite{he2016m2dp} solves the loop detection problem by projecting 
	the 3D point cloud to multiple 2D planes. Then the point cloud descriptor is obtained from the density signature for each of the 2D planes. Besides, Scan-Context \cite{scan-context} and LiDAR-Iris \cite{Wang2020LiDARIF} exploit the expanded bird view of the point cloud and describe the point cloud by encoding height information of the surrounding objects.

	\subsection{Learning Based Methods for PR and LCD}
	The first end-to-end work for large scale point cloud based place recognition is PointNetVLAD \cite{Uy2018PointNetVLADDP}. It treats PointNet \cite{pointnet} as the local feature extractor and aggregates the features with the NetVLAD \cite{NetVlad} pooling layer to produce a discriminative global descriptor. LPD-Net \cite{Liu-LPD-Net} extracts local contextual information using graph neural network inspired by DGCNN \cite{Wang2019DynamicGC} and intensively relies on additional handcrafted features, such as change of curvature and 2D scattering. PCAN \cite{Zhang2019PCAN3A} uses PointNet \cite{pointnet} to extract local point features, and then it makes use of PointNet++ \cite{Qi2017PointNetDH}  to produce an attention map which is integrated into the NetVLAD \cite{NetVlad} layer during feature aggregation.  
	Locus \cite{vid2021locus} extracts and encodes topological and temporal information related to components in a scene thus relying on the segmentation results.
	Different from the PointNet and graph convolution, Xia et al. \cite{Xia2020SOENetAS} designed the SOE-Net to extract local features, which introduces the PointOE module by integrating the local information from eight orientations inspired by PointSIFT \cite{PointSIFT}.
	DH3D \cite{du2020dh3d} recognizes the places and refines the 3D pose with a Siamese network simultaneously. LCDNet \cite{lcdnet} designs a network with a shared encoder and two heads used for global descriptors and relative pose estimation, respectively. Similar to LCDNet, OverlapNet \cite{chen2020overlapnet} also accomplishes the two tasks but assumes the relative pose is only up to a yaw angle.

	\subsection{Transformer Models in Point Cloud Processing}
	Guo et al. \cite{Guo2020PCTPC} proposes PCT (Point Cloud Transformer) for point cloud classification and segmentation, which stacks multiple self-attention modules and concatenates the features as contextual information. Point Transformer \cite{point_transformer} is proposed by Zhao et al. in a deep network architecture and achieves the state-of-the-art performance on classification and segmentation. As did in image-based transformer models, this method enlarges the receptive field by sampling and grouping layer by layer to capture the global information.
	In dealing with loop closure detection task, DAGC \cite{DAGC} introduces a dual attention mechanism consisting of a point-wise attention as well as a channel-wise attention and combines with graph convolution to describe a point cloud. NDT-Transformer \cite{Zhou2021NDTTransformerL3} and SOE-Net \cite{Xia2020SOENetAS} uses the handcrafted NDT cells and raw point data as the input of the network, respectively, followed by a point-wise transformer to build the global relations.

	\section{The Hierarchical Transformer Model for Large Scale Place Recognition}
	\label{sec:method}

	The self-attention in Transformer models is the key component used to extract more correlation information from the input data. The whole process involves linearly projecting each input embedding (i.e., the transformed data by linear operations) into three different spaces, producing three new representations known as $Query$, $Key$ and $Value$.
	These new embeddings will be used to obtain a score that can represent the dependency or correlation among the input points. A self-attention function can be described as mapping a $Query$ and a set of $Key$-$Value$ pairs to an output.
	
	In this section, we introduce the details of the proposed transformer model for place recognition based on the point cloud. Fig. \ref{fig:long-range-transformer} shows the overall  architecture of our proposed point cloud based hierarchical transformer network,
	where the short range transformer 
	aims at producing local descriptors, and the stacked long range transformer 
	concentrates on gathering global contextual information before aggregating a distinctive global descriptor. 
	
	
	For easier understanding of our work, by analogy with the original transformer in NLP \cite{tranformer}, we treat each point as a word and its neighbors as a sentence in the short range transformer. 
	The local feature of the point is represented with the sum of neighbors weighted by self-attention within the divided point cells.
	In the long range transformer, the whole point cloud is treated as a sentence, in which each point cell is a word. We will build the relations among all of the point cells in the point cloud.
	
	\subsection{The Point Cell Generation}
	\label{subsec:Cells}	
	We downsample from the original point cloud utilizing FPS (Farthest Point Sampling) algorithm \cite{Qi2017PointNetDH} to generate the point cells,
	
	\begin{equation}
		\label{equ:cells}
		C=[C_1,C_2,...,C_N] \in \mathbb{R}^{N\times (K+1)\times3},	
	\end{equation}	
	where $C_n=[P_n, P_{n1}, P_{n2},...,P_{nK}] \in \mathbb{R}^{(K+1)\times3}, n=1,2,...,N$ and $N$ is the number of the point cells in the sampled point cloud depending on the sampling rate $\tau$. The point cell $C_n$ contains $K+1$ points, with $P_{n}$ being the center point of it, and $P_{nk}$ being the neighboring point of $P_{n}$  searched from the original point cloud, where $k=1,2,...,K$. The benefits arising from the point cell generation include a limitation on the number of points involved in the computation of 
	transformer's features, as well as preserving the original point cloud's topological information as much as possible.
	The cells may overlap with each other especially when $K$ is larger than sampling rate $\tau$. The points in the cells are transformed to embedding vectors by the same MLP (Multi-Layer Perceptron) block, and the dimension of the transformed embeddings $C^{'}_n=[E_n, E_{n1}, E_{n2},...,E_{nK}]$ is $ \mathbb{R}^{(K+1)\times D_I}$.

	\subsection{The Short Range Transformer for Local Features}
	\label{subsec:SRT}

	In the short range transformer (SRT) block shown in Fig. \ref{fig:short-range-transformer}, we extract a local feature for each point cell. 
	As mentioned above, there are two types of point embedding in each point cell, i.e., one for the center point and the other for its neighbors. We use the terminologies $Query$ ($Q_S$), $Key$ ($K_S$) and $Value$ ($V_S$) as the input embedding for the short range transformer block, where $Q_S$ is an embedding which is transformed from the center-point embedding through a linear transformation, while $K_S$ and $V_S$ are obtained from the neighboring point embedding through two independent linear transformations, respectively. The local feature of each point cell is extracted by calculating a weighted sum of $V_S$, where the weights are obtained according to the relation between $Q_S$ and $K_S$.

	We use a subtraction operation \cite{Zhao2020ExploringSF} as the relation function to achieve the self-attention in our short range transformer block. Instead of adopting the dot product between $Q_S$ and $K_S$ in the classic relation function, subtraction self-attention is implemented by the subtraction $Q_S-K_S$. Using subtraction operation, $Q_S$ has to be repeated first so that it can be subtracted from $K_S$ and $V_S$. This operation leads to a \textbf{vector attention} which is adaptive to both different $V_S$ and also along the feature dimension.
	The transformer and subtraction self-attention are shown in Eq. \ref{equ:SRT}. However after MLP and linear transformation, the coordinate values of the points in Euclidean space are lost. Thus we exploit the positional encoding $\delta=\gamma(P_n-P_{nk})$ to encode the relative positional relation between the center point and its neighbors in the real world, where $\gamma$ is also an MLP.
    
	\begin{equation}
		\label{equ:SRT}
		\begin{aligned}	
			S_n = \sum_{k=1}^{K}Softmax(LN(MLP(Q_S-K_S+\delta)))\odot(V_S+\delta),
		\end{aligned}
	\end{equation}			
	where $LN$ is the Layer Normalization.
	
	To perform better, the position encoding is added to both self-attention and to $V_S$ as in \cite{point_transformer}. Through the SRT block, a sequence of local features $F_S$ corresponding to point cells $C$ are obtained as follows:

	\begin{equation}
		\label{equ:local_feat}
		F_S=[F_{S1},F_{S2},...,F_{SN}] \in \mathbb{R}^{N\times D_{S}},	
	\end{equation}	
	where $D_S$ is the dimension for each local feature.
	
	\subsection{The Long Range Transformer for Global Information}
	\label{subsec:LRT}
	Long range transformer (LRT) is the 
	the second stage of the point feature extraction procedure in our HiTPR network. This module can encode the global contextual relations among all of the point cells. Different from the previous transformer based methods which count in all points, we only exploit the local features extracted from the point cells, whose quantity is much smaller than that of the points in the original point cloud.
	In this stage, we take the local features $F_S=[F_{S1},F_{S2},...,F_{SN}]$ from the SRT block as the input of the LRT, and feed them into one MLP block, which is followed by three independent parallel linear transformations $W_Q$, $W_K$ and $W_V$ to produce the Query ($Q_L \in \mathbb{R}^{N\times D_k}$), Key ($K_L \in \mathbb{R}^{N\times D_k}$) and Value ($V_L \in \mathbb{R}^{N\times D_v}$) for the LRT module, shown in Fig.\ref{fig:long-range-transformer} and Eq.\ref{equ:long_qkv}.
		
	\begin{equation}
		\label{equ:long_qkv}
		Q_L,K_L,V_L=MLP(F_S)*(W_Q, W_K, W_V).
	\end{equation}
	Note that $W_q$, $W_k$ and $W_v$ are trained parameters as well. Afterwards, we utilize dot-product as the attention function and feed $Q_L$, $K_L$ and $V_L$ into the attention module to generate the result (Eq.\ref{equ:atte_long}),

	\begin{equation}
		\label{equ:atte_long}
		Attn=LRT(Q_L,K_L,V_L)=Softmax(\frac{Q_LK^T_L}{\sqrt{D_k})})V_L,  
	\end{equation}	
	where $D_k$ is the dimension of $K_L$.
	Throughout the linear transformation $Lin$, batch normalization $BatchNorm$, and activation function $Act$ in sequence, the result from attention module is transformed to produce the LRT feature with the residual connection (Eq.\ref{equ:long_feat}),

	\begin{equation}
		\label{equ:long_feat}
		F^m_L=Act(BatchNorm(Lin(Attn)))+F^{m-1}_L,
	\end{equation} 	
	where $m=1,2,...,M$ indicates the index of the stacked LRT blocks and $F^0_L=MLP(F_S)$. We concatenate the features from multiple LRT blocks along the dimension to generate the total long range feature (Eq. \ref{equ:stack_long_feat}),
	
	\begin{equation}
		\label{equ:stack_long_feat}
		F_{L}=Concat(F^m_L) \in \mathbb{R}^{N\times D_L}, 
	\end{equation}	
	where $D_L$ is the dimension of the long range feature, which is $M$ times the dimension of features extracted from a single LRT block.

    \subsection{Global Descriptor and Metric Learning}
	\label{subsec:aggregation}
	
	To represent the point cloud effectively, instead of adopting NetVLAD \cite{NetVlad} as in \cite{Liu-LPD-Net,Uy2018PointNetVLADDP}, we aggregate a global descriptor $F_G \in \mathbb{R}^{1\times D_G}$ by a max-pooling operator on the $F_{L} \in \mathbb{R}^{N\times D_L}$, because we have already encoded the global contextual information in long range transformer. 
	The aggregation module sequentially consists of a linear transformation, a batch normalization, an activation function as well as a max-pooling operator.
	
	At the end of the network, the loss function for metric learning is used for minimizing the Euclidean distance between descriptors extracted from point clouds sampled at the same place and maximizing the Euclidean distance between descriptors extracted from point clouds sample at different places. We use the lazy quadruplet loss proposed in \cite{Uy2018PointNetVLADDP} as the loss function, which can select hard pair samples to enhance the role of the negative samples on training. 
	\begin{table*}[h]
    	\caption{Comparison results on the average recall at top 1\% and at top 1 of different baseline networks trained on Ox. and tested on Ox., U.S., R.A. and B.D., respectively. HiTPR-S8 represents the results when sample rate is 8. The HiTPR-F8 represents the results to use the same handcrafted features as LPD-Net based on HiTPR-S8. The last two rows are used to separately illustrate the function of the handcrafted features and demonstrate our better results than the LPD-Net.} 
		\label{tab:baseline-result}
		\centering
		\begin{tabular}{c|c|c|c|c|c|c|c|c}  
			\hline       
			\hline
			\multirow{2}{15em}{\centering No Handcrafted Features} & \multicolumn{4}{c|}{Ave recall @ 1\%} & \multicolumn{4}{c}{Ave recall @ 1}\\    
			\cline{2-9} 
			& Ox. & U.S. & R.A. & B.D. & Ox. & U.S. & R.A. & B.D. \\
			\hline
			PN\_VLAD \cite{Uy2018PointNetVLADDP} &80.33&72.63&60.27&65.30&62.76&65.96&55.31&58.87\\
			\hline
			PCAN \cite{Zhang2019PCAN3A} &83.81&79.05&71.18&66.82&69.05&62.50&57.00&58.14\\
			\hline
			DH3D-4096 \cite{du2020dh3d} &84.26&-&-&-&73.28&-&-&-\\
			\hline
			DAGC \cite{DAGC}&87.49&83.49&75.68&71.21&73.34&-&-&-\\
			\hline
			LPD-Net* \cite{Liu-LPD-Net} &91.61&86.02&78.85&75.36&82.41&77.25&65.66&69.51 \\
			\hline
			
			SOE-Net* \cite{Xia2020SOENetAS}&93.41&-&-&-&84.20&-&-&-\\
			\hline
			
			\textbf{HiTPR-S8(Ours)} &93.61&85.57&75.00&71.30&86.61&75.78&63.10&64.45 \\ 
			\hline
			\textbf{HiTPR(Ours)} &\textbf{93.71}&\textbf{90.21}&\textbf{87.16}&\textbf{79.79}&\textbf{86.63}&\textbf{80.86}&\textbf{78.16}&\textbf{74.26} \\ 
			\hline
			\hline   
			\multirow{2}{15em}{\centering  Handcrafted Features} & \multicolumn{4}{c|}{Ave recall @ 1\%} & \multicolumn{4}{c}{Ave recall @ 1}\\   
			\cline{2-9} 
			& Ox. & U.S. & R.A. & B.D. & Ox. & U.S. & R.A. & B.D. \\
			\hline
			LPD-Net \cite{Liu-LPD-Net} &94.20&\textbf{94.71}&\textbf{89.65}&87.04&85.57&84.02&81.18&81.19\\
			\hline
			\textbf{HiTPR-F8(Ours)} &\textbf{94.64}&94.01&89.11&\textbf{88.31} & \textbf{87.77} &\textbf{86.00}&\textbf{81.32}&\textbf{81.80} \\ 
			\hline
			\hline
		\end{tabular}
	\end{table*}

	\section{Experiments}
	\label{sec:result}
	
	\subsection{Benchmark Datasets and Experimental Settings}
	\label{subsec:benchmark}
	\textbf{Datasets:} As in previous papers, we use the benchmark datasets created in \cite{Uy2018PointNetVLADDP} to train and evaluate our proposed model. It contains a modified Oxford RobotCar dataset and three in-house datasets: University Sector(U.S.), Residential Area(R.A.), and Business District(B.D.). The Oxford RobotCar dataset was collected within a pretty long period, capturing many different combinations of weather, traffic and pedestrians, including reverse loop, along with long-term changes such as construction and roadworks \cite{RobotCarDatasetIJRR}.
	Each point cloud in all of the datasets has 4096 points after removing the ground plane, with each point's coordinate value shifted to zero mean and re-scaled to the range of [-1,1].
	Two point clouds are defined as positive pairs if the distance between their centers is no larger than 10m, while two point clouds whose distance is over 50m are defined as negative pairs.
	For better training and evaluation, the Oxford datasets are divided into two parts, 21711 submaps for training and 3030 submaps for test, both of which do not overlap with each other. The other three inhouse datasets are also used for the benchmark datasets in the previous papers, which contain much fewer samples (about 4500 submaps totally) compared with the Ox. dataset. Because of the imbalance problem between the Ox. and inhouse datasets, we only used the latter for test, which are not used for training.
	\textbf{Evaluation metrics:} Similar to the previous works \cite{Uy2018PointNetVLADDP,Liu-LPD-Net,Zhang2019PCAN3A}, we adopt the top 1\% (@1\%) and top 1 (@1) average recall rate as the evaluation metrics. 
	\textbf{Implementation details:} Our HiTPR network consists of four modules, i.e., a point cell generator, a short range transformer (SRT), a stacked long range transformer (LRT), and a feature aggregate. In the point cell generation module, the sampling rate, $\tau$ and the number of neighboring points, $K$, are set to 4 and 32, respectively. The dimension of the input embedding, $D_I$, is 64. 
	In the SRT module, 
	the dimensions of $Q_S$, $K_S$, $V_S$ and the positional encoding $\delta$ are all set to 512. The dimension of the local features output from the SRT module is 64. In the LRT module, the dimension $D_k$ for $Q_L$ and $K_L$ is set to 64, while the dimension $D_v$ for $V_L$ is set to 256.
	The dimension of the feature output from a single LRT block is set to 256. Four LRT blocks are cascaded to construct the stacked long range transformer module, with the dimension $D_L$ of the concatenated long range features being 1024. 
	We feed the final long range feature into the aggregation module to generate the global feature whose dimension $D_G$ is 1024. At the end of the network, we use lazy quadruplet loss as the loss function and set the hyperparameters $\alpha$ and $\beta$ to 0.5 and 0.2, respectively. We use Adam \cite{kingma2014adam} with an initial learning rate of $5\times10^{-5}$ to train our network 20 epochs iteratively on the Pytorch platform. The learning rate decays to $1\times10^{-5}$ eventually. The numbers of positive samples and negative samples in the quadruplet are 2 and 8, respectively.
	
	\begin{figure*}[t]
		\centering
		\subfigure[Oxford]{
			\begin{minipage}[t]{0.25\textwidth}
				\centering
				\includegraphics[width=1\textwidth]{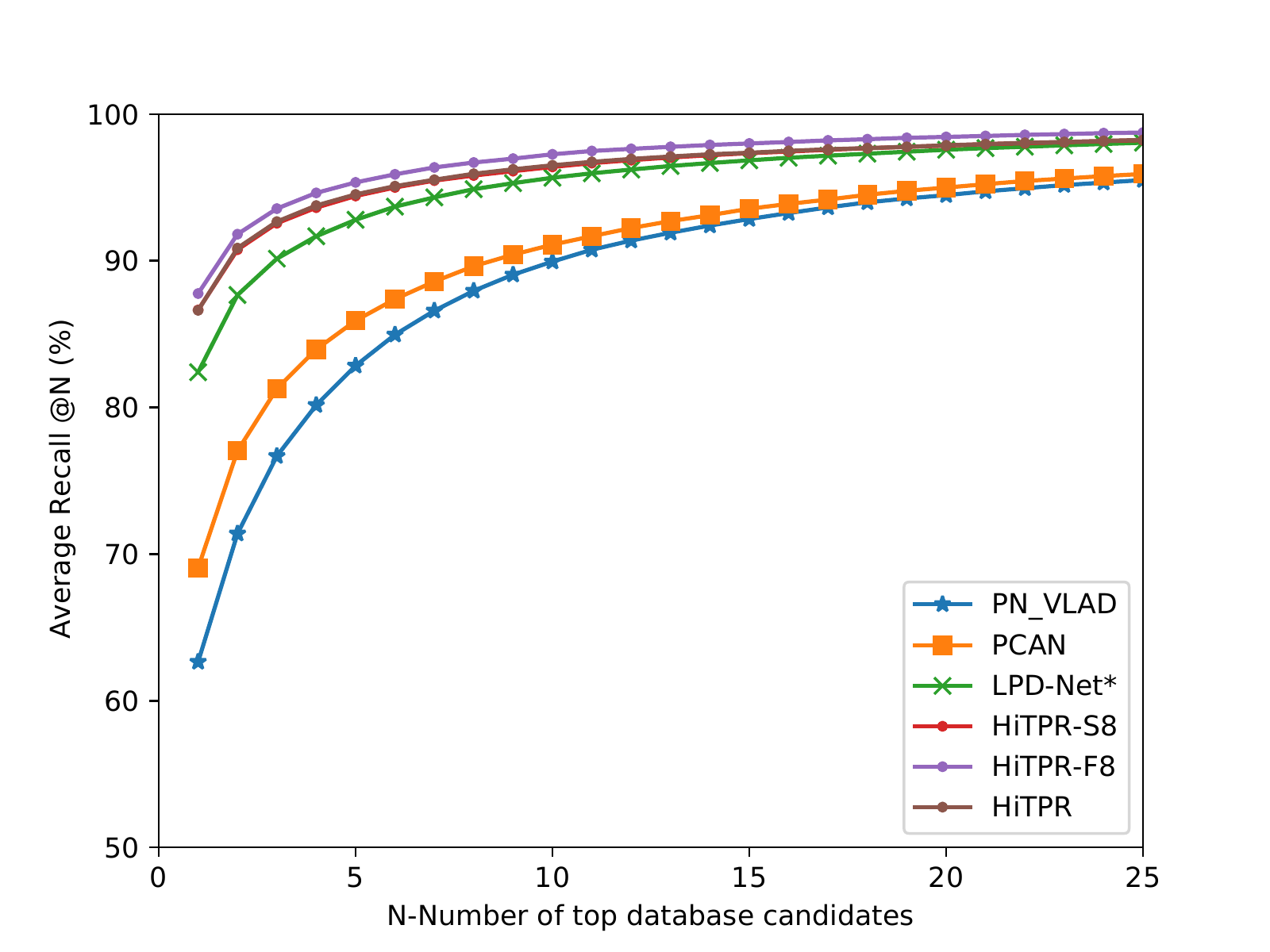}
			\end{minipage}%
		}%
		\subfigure[U.S.]{
			\begin{minipage}[t]{0.25\textwidth}
				\centering
				\includegraphics[width=1\textwidth]{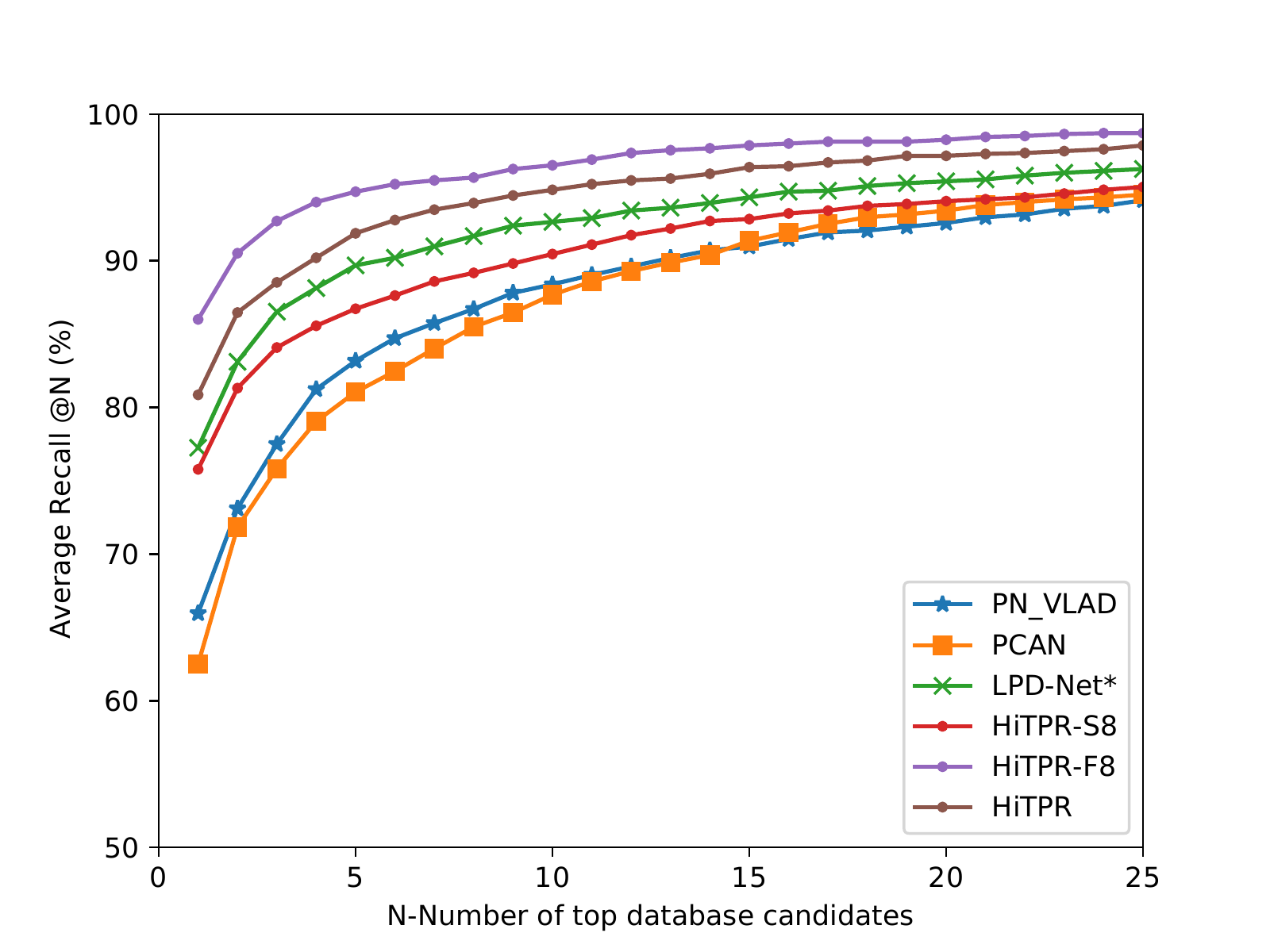}
			\end{minipage}%
		}%
		\subfigure[R.A.]{
			\begin{minipage}[t]{0.25\textwidth}
				\centering
				\includegraphics[width=1\textwidth]{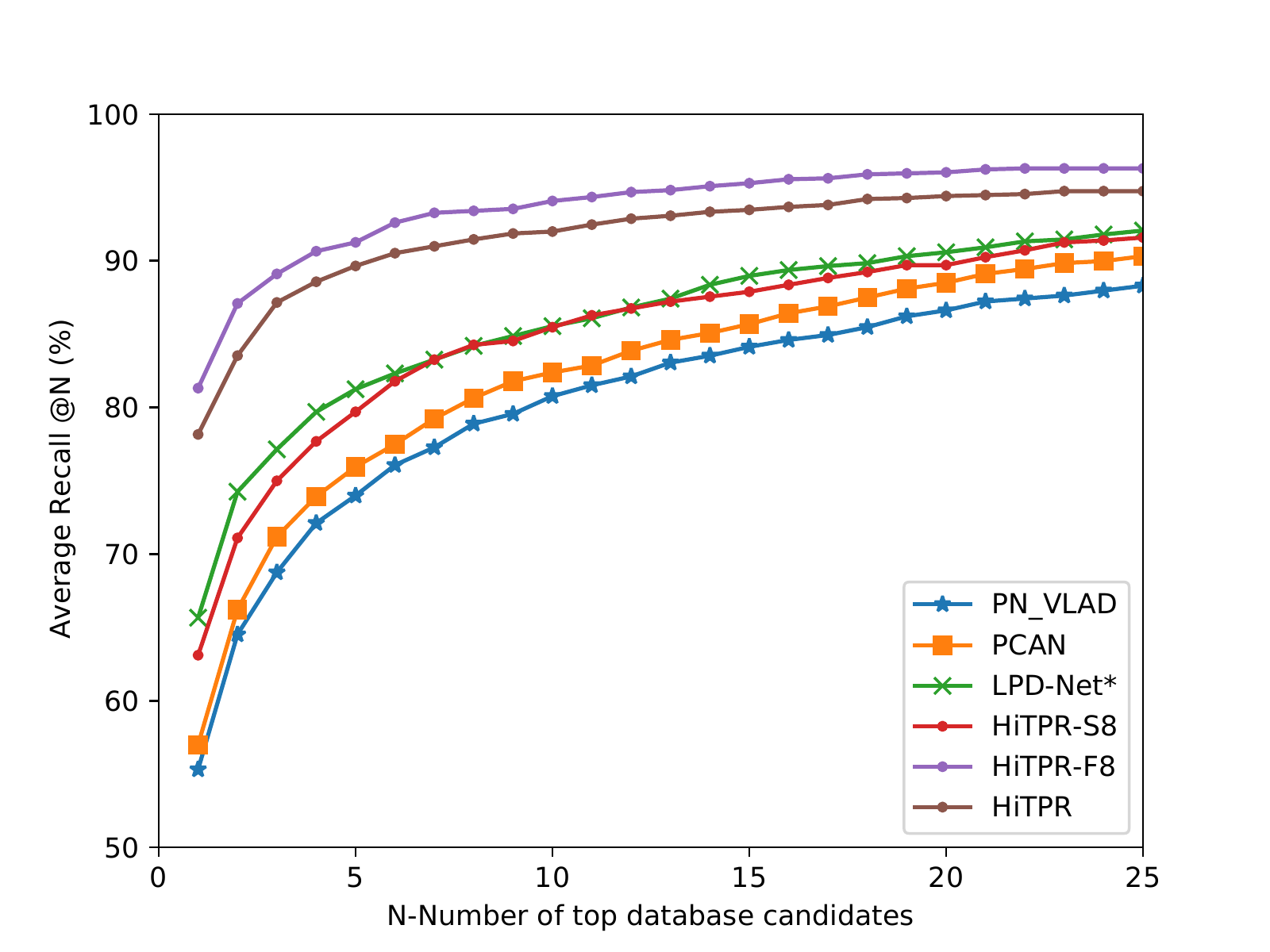}
			\end{minipage}%
		}%
		\subfigure[B.D.]{
			\begin{minipage}[t]{0.25\textwidth}
				\centering
				\includegraphics[width=1\textwidth]{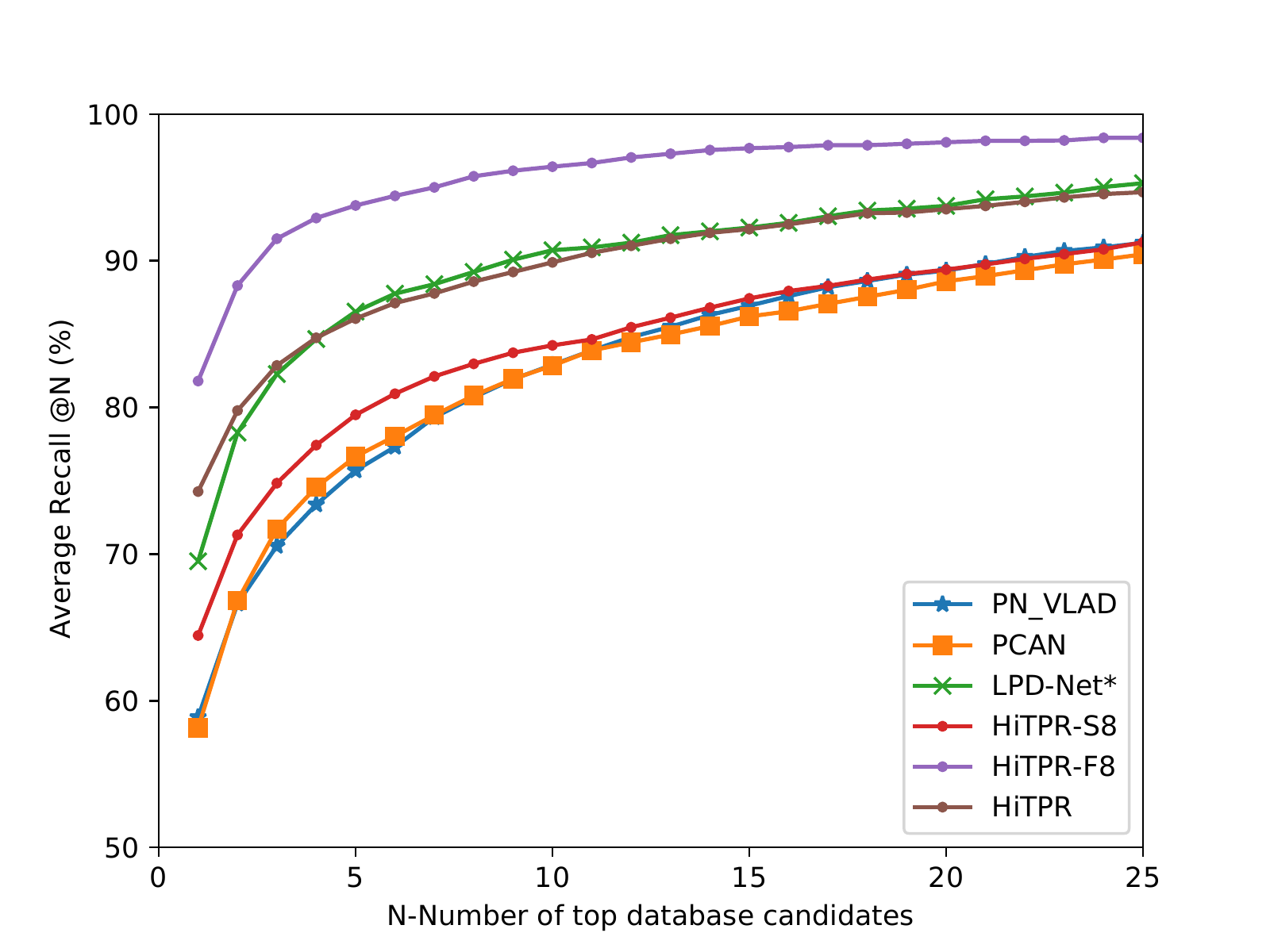}
			\end{minipage}%
		}%
		\centering
		\caption{Average recall of different baseline networks trained on the Oxford dataset, and tested on the Oxford, U.S, R.A. and B.D. dataset, respectively.}
		\label{fig:recall-curve}
	\end{figure*}

	\begin{table*}[h]
        \caption{Ablation studies of neighboring size $K$, dimension $D_S$, dimension $D_G$ and sample rate $\tau$}
        \label{tab:ablation_studies}
        \begin{center}
		\begin{tabular}{c|c|c||c|c|c||c|c|c||c|c|c}  
			\hline       
			\hline
			Neighbors& @ 1\%& @ 1 & Dimension $D_S$ & @ 1\% &  @ 1 & Dimension $D_G$&@ 1\%& @ 1 & Sampling rate $\tau$&@ 1\%& @ 1\\    
			\hline
			K = 8&90.40&80.73 &$D_S$ = 32&92.72&84.97 &$D_G$ = 32&83.45&68.48&$\tau$ = 1&93.14&86.04 \\
			\hline  
			K = 12&92.60&84.53 & $D_S$ = 64&\textbf{93.06}&\textbf{85.24}&$D_G$ = 64&86.18&73.25&$\tau$ = 2&\textbf{94.17}&\textbf{87.91} \\
			\hline  
			K = 16&92.62&85.15&$D_S$ = 128&91.66&83.15&$D_G$ = 128&89.10&78.26&$\tau$ = 4&93.06&85.24 \\
			\hline  
			K = 24&92.96&85.50 &$D_S$ = 256&91.29&82.95&$D_G$ = 256&91.26&82.09&$\tau$ = 8&92.62&85.15  \\		 
			\hline
			K = 28&93.53&86.36& $D_S$ = 512&92.50&84.85 & $D_G$ = 512&91.80&83.56&$\tau$ = 16&91.10&81.60\\		 
			\hline
			K = 32&\textbf{93.61}&\textbf{86.61} &$D_S$ = 1024&91.84&83.36&$D_G$ = 1024&\textbf{93.06}&\textbf{85.24}&$\tau$ = 32&88.70&76.74\\	
			\hline	 
			\hline
		\end{tabular}
        \end{center}
    \end{table*}
    
	\subsection{Quantitative Results}
	\label{subsec:QuanRes}

	The results of the compared methods are shown in Table \ref{tab:baseline-result}. It shows that our method HiTPR achieves best performance in terms of both distinctiveness and generalization, while the HiTPR-S8 represents the results when the sample rate $\tau$ is 8 for computation reduction purposes. HiTPR-F8 represents the results to use the same handcrafted features as LPD-Net based on HiTPR-S8.
	The LPD-Net* indicates the version in which the handcrafted features are removed from the original LPD-Net \cite{Liu-LPD-Net}. We remove the handcrafted features based on the facts that most previous works in LiDAR-based place recognition do not extract handcrafted features as their network input. Therefore, for a fair comparison, we also compare the LPD-Net without the handcrafted features, i.e., the LPD-Net*. With the handcrafted features acquired from offline pre-computation, we still use LPD-Net as the original one. The LPD-Net can achieve 94.20\% on top 1\% recall and 85.57\% on top 1 recall. Our HiTPR network can achieve a comparable performance with LPD-Net and even surpass LPD-Net at top 1 recall evaluation metric without the pre-computed handcrafted features. Our HiTPR-F8 can generally achieve better results than LPD-net. 
	The SOE-Net \cite{Xia2020SOENetAS} utilizes an improved metric learning loss function that plays the major role on the improvement. To fully reflect the feature extraction capability of our  model, and ignore the effect brought by different loss functions, we compared the result of the SOE-Net trained under lazy quadruplet loss function as used by the other compared works, which is named as SOE-Net* in Table \ref{tab:baseline-result}. In this paper, we focus on the methods applied in the SLAM system so that some transformer-based works relying on pre-processed Lidar database or queries, such as NDT-Transformer \cite{Zhou2021NDTTransformerL3}, are not compared here. Most of the existing works\cite{Xia2020SOENetAS, Zhou2021NDTTransformerL3} use transformer to extract global contextual features and then work along with NetVlad \cite{NetVlad}. Thus, transformer is only a part of their networks. In contrast, we extract point features with the transformer and propose an integral and hierarchical transformer-based network to extract the local and global features, where transformer acts as the core. 

	In addition, we show the average recall of different networks at top 25 on the Ox., U.S., R.A. and B.D. datasets in Fig. \ref{fig:recall-curve}. Obviously, our method can achieve more accurate results than the other compared methods, especially at top 1, which is more practical in real world applications.

    \subsection{Ablation Studies}
    \label{subsec:ablation}
	
	We conduct all the ablation studies including training and testing only on the Oxford RobotCar dataset. In the ablation studies for neighboring size $K$, sampling rate $\tau$ is set to 8, while in the ablation studies for $D_S$ and $D_G$, it is set to 4. The detailed results consisting of neighboring size $K$, dimension $D_S$ of SRT features, dimension $D_G$ of output features and sample rate $\tau$ are shown in Table \ref{tab:ablation_studies}.

	\textbf{Different neighboring size $K$ for SRT:} We implement the short range transformer with different number $K$ of neighboring points, which affects the size of point cells. 
	It demonstrates that the recall rate improves as $K$ increases. However, considering the time cost and memory consumption, we do not keep increasing and only select an acceptable value to complete our experiments described above. For memory reduction purposes, we set $K=16$ other than $K=32$ to implement the subsequent ablation experiments.

	\textbf{Different dimension $D_S$ of SRT features:} 
	The study demonstrates that our network can achieve the best performance when $D_S$ equals 64. This setting is preserved in the other ablation studies.

	\textbf{Different dimension $D_G$ of the output features:} The ablation experiment on $D_G$ shows that the global descriptors have more expressive ability with higher dimension.
	We choose $D_G=1024$ as our experimental setting. Although the dimension used in our work is higher than the previous papers, they achieve better or similar results in the lower dimension, e.g. the 1\% recall is 94.92 and 94.66 when the dimension is 256 and 512 respectively in LPD-Net\cite{Liu-LPD-Net}.
	
	\textbf{Different sampling rate $\tau$:} We set different $\tau$ to implement this ablation experiment. 
	Table \ref{tab:ablation_studies} demonstrates that when the sample rate is relatively larger, $\tau=$ 16 and 32, the recall increases significantly as the sample rate decreases. When the sample rate is small such as $\tau=$ 2, 4 and 8, the recall has small changes, and even when we use more point cells, e.g., $\tau$ = 1, the recall has no further increase.
	Therefore, our approach can produce a robust global descriptor even though the downsampled point cloud is extremely sparse for computation and memory reduction purposes.

	
	\begin{table}[h]
        \caption{Memory and computation requirements.}
        \label{tab:computational cost}
        \begin{center}	
	
		\begin{tabular}{c|c|c|c}  
	        \hline 
			\hline
			Network&Params&FLOPs&Runtime\\    
			\hline
			PN\_VLAD\cite{Uy2018PointNetVLADDP}&19.78M&214&37.8ms \\	
			\hline	 
		    PCAN\cite{Zhang2019PCAN3A}&20.42M&339&102.1ms\\
		    \hline
		    LPD-Net\cite{Liu-LPD-Net}&19.81M&353&49.3ms\\
		    \hline
		    HiTPR(Ours)&2.72M&308&36.2ms\\
			\hline	 
			\hline
		\end{tabular}
        \end{center}
    \end{table}    
	
	\textbf{Computational Cost Analysis}:
	We evaluate required computational resources for different networks in terms of the number of parameters, floating point operations (FLOPs) and runtime per frame. All networks are run on a single NVIDIA Tesla P40 GPU (24G memory). We present the results in Table \ref{tab:computational cost}.  Compared with the previous works, our network costs shorter time per frame with fewer parameters.

	
	\section{Conclusion}
	\label{sec:conclusion}
	In this paper, we propose a hierarchical transformer network to extract a discriminative and generalized descriptor from point cloud for place recognition. Our approach leverages short range transformer to encode the local feature of each point cell and long range transformer to gather contextual information for all of the point cells. Experiments on standard benchmark datasets demonstrate the superiority and effectiveness of our approach.
	
	
	\noindent\textbf{Acknowledgement}: This paper is supported by National Key Research and Development Program of China (No. 2019YFB2102100), the Science and Technology Development Fund of Macau SAR (File no. 0015/2019/AKP), Guangdong-Hong Kong-Macao Joint Laboratory of Human-Machine Intelligence-Synergy Systems (No. 2019B121205007), and the National Natural Science Foundation of China (No. 61803083). 

	%
	
	
	
	\clearpage 
	
%
	
	\bibliographystyle{IEEEtran}
	\bibliography{example}  

\begin{thebibliography}{10}
\providecommand{\url}[1]{#1}
\csname url@rmstyle\endcsname
\providecommand{\newblock}{\relax}
\providecommand{\bibinfo}[2]{#2}
\providecommand\BIBentrySTDinterwordspacing{\spaceskip=0pt\relax}
\providecommand\BIBentryALTinterwordstretchfactor{4}
\providecommand\BIBentryALTinterwordspacing{\spaceskip=\fontdimen2\font plus
\BIBentryALTinterwordstretchfactor\fontdimen3\font minus
  \fontdimen4\font\relax}
\providecommand\BIBforeignlanguage[2]{{%
\expandafter\ifx\csname l@#1\endcsname\relax
\typeout{** WARNING: IEEEtran.bst: No hyphenation pattern has been}%
\typeout{** loaded for the language `#1'. Using the pattern for}%
\typeout{** the default language instead.}%
\else
\language=\csname l@#1\endcsname
\fi
#2}}

\bibitem{VPR-Bench}
Z.~{Mubariz}, S.~{Garg}, M.~{Milford}, J.~{Kooij}, D.~{Flynn},
  K.~{McDonald-Maier}, and S.~{Ehsan}, ``Vpr-bench: An open-source visual place
  recognition evaluation framework with quantifiable viewpoint and appearance
  change,'' in \emph{IJCV}, 2021.

\bibitem{Wang2020LiDARIF}
Y.~Wang, Z.~Sun, C.~Xu, S.~Sarma, J.~Yang, and H.~Kong, ``Lidar iris for
  loop-closure detection,'' \emph{IROS}, pp. 5769--5775, 2020.

\bibitem{segmatch2017}
R.~Dub{\'e}, D.~Dugas, E.~Stumm, J.~Nieto, R.~Siegwart, and C.~Cadena,
  ``Segmatch: Segment based place recognition in 3d point clouds,'' in
  \emph{ICRA}, 2017, pp. 5266--5272.

\bibitem{Uy2018PointNetVLADDP}
M.~A. Uy and G.~H. Lee, ``Pointnetvlad: Deep point cloud based retrieval for
  large-scale place recognition,'' \emph{CVPR}, pp. 4470--4479, 2018.

\bibitem{Liu-LPD-Net}
Z.~Liu, S.~Zhou, C.~Suo, P.~Yin, W.~Chen, H.~Wang, H.~Li, and Y.~Liu,
  ``Lpd-net: 3d point cloud learning for large-scale place recognition and
  environment analysis,'' in \emph{ICCV}, 2019, pp. 2831--2840.

\bibitem{du2020dh3d}
J.~Du, R.~Wang, and D.~Cremers, ``Dh3d: Deep hierarchical 3d descriptors for
  robust large-scale 6dof relocalization,'' in \emph{ECCV}, 2020.

\bibitem{vid2021locus}
K.~Vidanapathirana, P.~Moghadam, B.~Harwood, M.~Zhao, S.~Sridharan, and
  C.~Fookes, ``Locus: Lidar-based place recognition using spatiotemporal
  higher-order pooling,'' in \emph{ICRA}, 2021.

\bibitem{Xia2020SOENetAS}
Y.~Xia, Y.~Xu, S.~Li, R.~Wang, J.~Du, D.~Cremers, and U.~Stilla, ``Soe-net: A
  self-attention and orientation encoding network for point cloud based place
  recognition,'' \emph{CVPR}, 2021.

\bibitem{Zhou2021NDTTransformerL3}
Z.~Zhou, C.~Zhao, D.~Adolfsson, S.~Su, Y.~Gao, T.~Duckett, and L.~Sun,
  ``Ndt-transformer: Large-scale 3d point cloud localisation using the normal
  distribution transform representation,'' \emph{ICRA}, 2021.

\bibitem{lcdnet}
D.~Cattaneo, M.~Vaghi, and A.~Valada, ``Lcdnet: Deep loop closure detection for
  lidar slam based on unbalanced optimal transport,''
  \emph{arXiv:2103.05056v1}, 2021.

\bibitem{Komorowski_2021_WACV}
J.~Komorowski, ``Minkloc3d: Point cloud based large-scale place recognition,''
  in \emph{WACV}, January 2021, pp. 1790--1799.

\bibitem{chen2020overlapnet}
X.~Chen, T.~L{\"a}be, A.~Milioto, T.~R{\"o}hling, O.~Vysotska, A.~Haag,
  J.~Behley, C.~Stachniss, and F.~Fraunhofer, ``Overlapnet: Loop closing for
  lidar-based slam,'' in \emph{Proc. of Robotics: Science and Systems (RSS)},
  2020.

\bibitem{he2016m2dp}
L.~He, X.~Wang, and H.~Zhang, ``M2dp: A novel 3d point cloud descriptor and its
  application in loop closure detection,'' in \emph{IROS}, 2016, pp. 231--237.

\bibitem{kim2018scan}
G.~Kim and A.~Kim, ``Scan context: Egocentric spatial descriptor for place
  recognition within 3d point cloud map,'' in \emph{IROS}, 2018, pp.
  4802--4809.

\bibitem{Zhang2019PCAN3A}
W.~Zhang and C.~Xiao, ``Pcan: 3d attention map learning using contextual
  information for point cloud based retrieval,'' \emph{CVPR}, pp.
  12\,428--12\,437, 2019.

\bibitem{NetVlad}
R.~{Arandjelovic}, P.~{Gronat}, A.~{Torii}, T.~{Pajdla}, and J.~{Sivic},
  ``Netvlad: Cnn architecture for weakly supervised place recognition,'' in
  \emph{CVPR}, 2016, pp. 5297--5307.

\bibitem{tranformer}
A.~Vaswani, N.~Shazeer, N.~Parmar, J.~Uszkoreit, L.~Jones, A.~N. Gomez,
  u.~Kaiser, and I.~Polosukhin, ``Attention is all you need,'' in
  \emph{NeurIPS}, 2017, p. 6000–6010.

\bibitem{devlin-etal-2019-bert}
J.~Devlin, M.-W. Chang, K.~Lee, and K.~Toutanova, ``Bert: Pre-training of deep
  bidirectional transformers for language understanding,'' in \emph{ACL}, 2019,
  pp. 4171--4186.

\bibitem{brown2020language}
T.~Brown, B.~Mann, N.~Ryder, M.~Subbiah, J.~D. Kaplan, P.~Dhariwal,
  A.~Neelakantan, P.~Shyam, G.~Sastry, A.~Askell, S.~Agarwal, A.~Herbert-Voss,
  G.~Krueger, T.~Henighan, R.~Child, A.~Ramesh, D.~Ziegler, J.~Wu, C.~Winter,
  C.~Hesse, M.~Chen, E.~Sigler, M.~Litwin, S.~Gray, B.~Chess, J.~Clark,
  C.~Berner, S.~McCandlish, A.~Radford, I.~Sutskever, and D.~Amodei, ``Language
  models are few-shot learners,'' in \emph{NeurIPS}, vol.~33, 2020, pp.
  1877--1901.

\bibitem{chen2020generative}
M.~Chen, A.~Radford, R.~Child, J.~Wu, H.~Jun, D.~Luan, and I.~Sutskever,
  ``Generative pretraining from pixels,'' in \emph{ICML}, 2020, pp. 1691--1703.

\bibitem{detr}
N.~Carion, F.~Massa, G.~Synnaeve, N.~Usunier, A.~Kirillov, and S.~Zagoruyko,
  ``End-to-end object detection with transformers,'' in \emph{ECCV}, 2020, pp.
  213--229.

\bibitem{dosovitskiy2020image}
A.~Dosovitskiy, L.~Beyer, A.~Kolesnikov, D.~Weissenborn, X.~Zhai,
  T.~Unterthiner, M.~Dehghani, M.~Minderer, G.~Heigold, S.~Gelly,
  \emph{et~al.}, ``An image is worth 16x16 words: Transformers for image
  recognition at scale,'' in \emph{ICLR}, 2020.

\bibitem{ipt}
H.~Chen, Y.~Wang, T.~Guo, C.~Xu, Y.~Deng, Z.~Liu, S.~Ma, C.~Xu, C.~Xu, and
  W.~Gao, ``Pre-trained image processing transformer,'' in \emph{CVPR}, 2021,
  pp. 12\,299--12\,310.

\bibitem{ESF}
W.~Wohlkinger and M.~Vincze, ``Ensemble of shape functions for 3d object
  classification,'' in \emph{ROBIO}, 2011, pp. 2987--2992.

\bibitem{point-hist}
R.~B. Rusu, N.~Blodow, Z.~C. Marton, and M.~Beetz, ``Aligning point cloud views
  using persistent feature histograms,'' in \emph{IROS}, 2008, pp. 3384--3391.

\bibitem{fpfh}
R.~B. Rusu, N.~Blodow, and M.~Beetz, ``Fast point feature histograms (fpfh) for
  3d registration,'' in \emph{ICRA}, 2009, pp. 3212--3217.

\bibitem{scan-context}
G.~Kim and A.~Kim, ``Scan context: Egocentric spatial descriptor for place
  recognition within {3D} point cloud map,'' in \emph{IROS}, 2018.

\bibitem{pointnet}
R.~Q. {Charles}, H.~{Su}, M.~{Kaichun}, and L.~J. {Guibas}, ``Pointnet: Deep
  learning on point sets for 3d classification and segmentation,'' in
  \emph{CVPR}, 2017, pp. 77--85.

\bibitem{Wang2019DynamicGC}
Y.~Wang, Y.~Sun, Z.~Liu, S.~E. Sarma, M.~Bronstein, and J.~Solomon, ``Dynamic
  graph cnn for learning on point clouds,'' \emph{ACM Transactions on Graphics
  (TOG)}, vol.~38, pp. 1 -- 12, 2019.

\bibitem{Qi2017PointNetDH}
C.~Qi, L.~Yi, H.~Su, and L.~Guibas, ``Pointnet++: Deep hierarchical feature
  learning on point sets in a metric space,'' in \emph{NeurIPS}, 2017.

\bibitem{PointSIFT}
M.~Jiang, Y.~Wu, T.~Zhao, Z.~Zhao, and C.~Lu, ``Pointsift: A sift-like network
  module for 3d point cloud semantic segmentation,'' \emph{arXiv preprint
  arXiv:1807.00652}, 2018.

\bibitem{Guo2020PCTPC}
M.~Guo, J.-X. Cai, Z.-N. Liu, T.-J. Mu, R.~Martin, and S.~Hu, ``Pct: Point
  cloud transformer,'' \emph{Computational Visual Media}, pp. 187--199, 2021.

\bibitem{point_transformer}
H.~Zhao, L.~Jiang, J.~Jia, P.~Torr, and V.~Koltun, ``Point transformer,'' in
  \emph{ICCV}, 2021.

\bibitem{DAGC}
Q.~Sun, H.~Liu, J.~He, Z.~Fan, and X.~Du, ``Dagc: Employing dual attention and
  graph convolution for point cloud based place recognition,'' in \emph{ICMR},
  2020, p. 224–232.

\bibitem{Zhao2020ExploringSF}
H.~Zhao, J.~Jia, and V.~Koltun, ``Exploring self-attention for image
  recognition,'' \emph{CVPR}, pp. 10\,073--10\,082, 2020.

\bibitem{RobotCarDatasetIJRR}
W.~Maddern, G.~Pascoe, C.~Linegar, and P.~Newman, ``{1 Year, 1000km: The Oxford
  RobotCar Dataset},'' \emph{IJRR}, vol.~36, no.~1, pp. 3--15, 2017.

\bibitem{kingma2014adam}
D.~P. Kingma and J.~Ba, ``Adam: A method for stochastic optimization,'' in
  \emph{ICLR}, 2015.

\end{thebibliography}
	
\end{document}